\documentclass[11pt]{article}
\usepackage{authblk}

\newlength{\defbaselineskip}
\usepackage{amsmath}
\usepackage{amsthm} 
\usepackage{amssymb}
\usepackage{stmaryrd}

\DeclareMathOperator*{\argmin}{arg\,min}
\DeclareMathOperator{\diag}{diag}

\newcommand{\defeq}{\stackrel{\text{\tiny def}}{=}}
\DeclareMathOperator{\proj}{\mathcal{P}}
\DeclareMathOperator{\prox}{prox}
\newcommand{\proxx}[1]{\prox_{#1 \|\cdot\|_1}}

\newcommand{\order}{\mathcal{O}}              
\newcommand{\x}{\mathbf{x}}
\newcommand{\e}{\mathbf{e}}
\newcommand{\z}{\mathbf{z}}
\renewcommand{\xi}{{\x}_{i}}
\newcommand{\X}{\mathbf{X}}

\newcommand{\y}{\mathbf{y}}

\newcommand{\W}{\mathbf{W}}
\newcommand{\R}{\mathbb{R}}    

\newcommand{\UU}{\mathbf{U}}

\newcommand{\MU}{\boldsymbol{\mu}}

\newcommand{\SSS}{\mathcal{S}}
\newcommand{\vv}{\mathbf{v}}
\newcommand{\CC}{\mathbf{C}}

\renewcommand{\c}{\mathbf{c}}
\renewcommand{\d}{\mathbf{d}}
\newcommand{\sss}{\mathcal{S}} 
\newcommand\restr[2]{#1_{\hspace{.3ex}\rule{.01pt}{.7em}\hspace{.3ex}#2}}
\newcommand{\Ast}{\star} 

\newcommand{\eye}{\mathbf{I}}

\newcommand{\ones}{\mathbbb 1}
\newcommand{\sparsity}{k} 

\DeclareMathAlphabet{\mathbbb}{U}{bbold}{m}{n}

\theoremstyle{plain}
\newtheorem{thm}{\protect\theoremname}
\theoremstyle{plain}

\providecommand{\lemmaname}{Lemma}
\providecommand{\theoremname}{Theorem}
\newtheorem{prop}[thm]{Proposition}

\newtheorem{remark}[thm]{Remark}

\usepackage{graphicx}
\usepackage[caption=false,font=footnotesize]{subfig}
\usepackage{url}
\usepackage{pifont}
\usepackage{multirow}

\usepackage{algorithm}
\usepackage[noend]{algpseudocode} 
\usepackage{enumitem}
\usepackage{booktabs}
\usepackage{siunitx}

\usepackage{xcolor}

\date{\vspace{-5ex}}

\begin{document}

\title{
Efficient Solvers for Sparse Subspace Clustering \footnote{This paper is accepted for publication in Signal Processing.}
}

\author{Farhad Pourkamali-Anaraki\\Department of Computer Science, University of Massachusetts Lowell, MA, USA \and James Folberth\\Department of Applied Mathematics, University of Colorado Boulder, CO, USA \and Stephen Becker \\Department of Applied Mathematics, University of Colorado Boulder, CO, USA }


\maketitle

\begin{abstract}
	Sparse subspace clustering (SSC)  clusters $n$  points that lie near a union of low-dimensional subspaces. The SSC model expresses each point as a linear or affine combination of the other points, using either $\ell_1$ or $\ell_0$ regularization. Using $\ell_1$ regularization results in a convex problem but requires $\order(n^2)$ storage, and is typically solved by the alternating direction method of multipliers which takes $\order(n^3)$ flops. The $\ell_0$ model is non-convex but only needs memory linear in $n$, and is solved via orthogonal matching pursuit and cannot handle the case of affine subspaces. This paper shows that a proximal gradient framework can solve SSC, covering both $\ell_1$ and $\ell_0$  models, and both linear and affine constraints. For both $\ell_1$ and $\ell_0$, 
algorithms to compute the proximity operator in the presence of affine constraints have not been presented in the SSC literature, 
so we derive an exact and efficient algorithm that solves the $\ell_1$ case with just $\order(n^2)$ flops. In the $\ell_0$ case, our algorithm retains the low-memory overhead, and is the first algorithm to solve the SSC-$\ell_0$ model with affine constraints. Experiments show our algorithms do not rely on sensitive regularization parameters, and they are less sensitive to sparsity misspecification and high noise.
\end{abstract}

\section{Introduction} \label{sec:intro}
In modern data analysis, clustering is an important tool for extracting information from large-scale data sets by identifying groups of similar data points without the presence of ground-truth labels. Therefore, there has been growing interest in developing accurate and efficient clustering algorithms by taking account of the intrinsic structure of large high-dimensional data sets. For instance, the popular K-means algorithm and its kernel-based variants are based on the assumption that
(mapped) data points are evenly distributed within linearly separable clusters \cite{pourkamali2017preconditioned,bachem2018scalable}.

There has been much work on approaches for more complicated clustering models, such as data that comes from a mixture of manifolds.
For some problems, a
reasonable  assumption is that of data points lying near a union of low-dimensional subspaces \cite{soltanolkotabi2012geometric,vidal2016sparse}. The dimensions and orientations of the subspaces are unknown and there are possibly non-trivial intersections between every pair of subspaces. The main task is to partition a given data set such that each group contains only data points from the same subspace. This problem is referred to as ``subspace clustering'' and has numerous applications in machine learning and computer vision such as motion segmentation and face clustering \cite{elhamifar2013sparse,heckel2017dimensionality}.

Among existing subspace clustering techniques, a popular line of work is focused on applying spectral clustering to an affinity matrix obtained by solving a global optimization problem, which represents each data point as a linear or affine combination of other points \cite{elhamifar2009sparse}. Given $\x_1,\ldots,\x_n$ that lie near a union of subspaces in $\R^p$, let $\X\in\R^{p\times n}$ be the matrix whose columns are the points. Then, each $\x_j$, $j=1,\ldots,n$, can be expressed as:
\begin{equation}
\x_j=\X\c_j+\e_j,\;\text{s.t.}\;[\c_j]_j=0,\;\c_j^T\ones=1,\label{eq:selfexp}
\end{equation}
where $\c_j\in\R^n$ is the coefficient vector and $\e_j\in\R^p$ is the representation error. The constraint $[\c_j]_j=0$ eliminates the trivial solution of expressing a point as a linear combination of itself. Also, the  constraint $\c_j^T\ones=1$ allows us to represent data points that lie near a union of affine rather than linear subspaces \cite{elhamifar2013sparse,patel2013latent,li2018geometric}.

When representing each data point in a low-dimensional subspace in terms of other points in the same subspace, the vector $\c_j$ in Eq.~\eqref{eq:selfexp} is not unique. However, the main goal is to find a ``subspace-preserving'' solution such that there are no connections between points from different subspaces. Thus,  $[\c_j]_i\neq 0$ should indicate that $\x_i$ is in the same subspace as $\x_j$. Given subspace-preserving representations $\CC=[\c_1,\ldots,\c_n]\in\R^{n\times n}$, a graph with $n$ vertices corresponding to data points is constructed where its affinity matrix is given by the symmetric  matrix $\W=|\CC|+|\CC^T|$. Then, spectral clustering \cite{von2007tutoria} is applied to $\W$ to cluster the data. Note that the subspace clustering problem  we consider in this work can be viewed as a particular case of spectral clustering in which the affinity matrix  is formed using the self-expressiveness property. More generally, spectral clustering uses manifold structures of data points and a distance metric for constructing graphs that represent such relationships \cite{schiebinger2015geometry,arias2019unconstrained,tremblay2020approximating}.

Sparse subspace clustering (SSC) approaches the problem of finding subspace-preserving coefficients by enforcing a sparsity prior on the columns of the matrix $\CC$. To do so, a popular technique  is centered on solving the following convex optimization program  \cite{elhamifar2009sparse,elhamifar2013sparse} (referred to as SSC-$\ell_1$ in this paper):
\begin{equation}
\min_{\CC}\|\CC\|_1+\frac{\lambda_e}{2}\|\X-\X\CC\|_F^2\;\text{s.t.}\;\diag(\CC)=\mathbf{0},\;\CC^T\ones=\ones,\label{eq:SSC_l1}
\end{equation}
where the $\ell_1$ norm promotes the sparsity of $\CC$ and $\lambda_e>0$ is the regularization parameter. Prior work has shown that the solution of \eqref{eq:SSC_l1} is guaranteed to be subspace-preserving under broad  conditions on the subspaces as well as under the presence of noise and outliers \cite{soltanolkotabi2014robust,you2015geometric,tsakiris2018theoretical}. Although SSC-$\ell_1$ is supported by a rich body of theory, the computational complexity associated with solving \eqref{eq:SSC_l1} using the alternating direction method of multipliers (ADMM, cf.\ \cite{boyd2011distributed}) scales cubically with the number of data points. In addition, the process of optimal parameter selection for ADMM requires a significantly increased amount of computational time \cite{xu2017admm}. In fact, as we will corroborate later, a poor parameter selection for ADMM leads to low  accuracy clustering results. Moreover, variants of ADMM with adaptive schemes for updating the solver parameter do not seem to be effective.

Therefore, despite the existence of strong theoretical guarantees, finding subspace-preserving coefficients based on $\ell_1$ norm regularization is computationally prohibitive for large-scale data sets \cite{adler2015linear,traganitis2017sketched,you2018scalable,abdolali2019scalable,pourkamali2019large}. One solution to this problem has been to 
use $\ell_0$ instead of $\ell_1$ regularization on the columns of $\CC$ \cite{dyer2013greedy,chen2018active}. The resulting model is the following non-convex optimization program (referred to as SSC-$\ell_0$ in this paper): for all $j=1,\ldots,n$, solve:
\begin{align}
\min_{\c_j} \frac{1}{2}\|\x_j-\X\c_j\|_2^2\;\text{s.t.}\;\|\c_j\|_0\leq \sparsity,\;[\c_j]_j=0,\;\c_j^T\ones=1.\label{eq:SSCl0}
\end{align}
If we remove the linear equality constraint $\c_j^T\ones=1$ associated with affine subspaces, then the $k$-sparse coefficient vector $\c_j$ can be estimated using the orthogonal matching pursuit (OMP) algorithm.
However, OMP cannot directly deal with the more general class of affine subspaces,
as OMP is a specialized greedy algorithm that enforces the sparsity constraint by only taking $k$ steps and cannot enforce any other kind of constraint. It is also worth pointing out that OMP is only known to solve the problem accurately under certain assumptions that do not hold in the subspace clustering problem.
In particular, the data matrix does not satisfy mutual incoherence or restricted isometry properties under the union of subspaces model. The work of \cite{you2016scalable} presents a theoretical analysis of the sparse subspace clustering problem using $\ell_0$ norm regularization for the noiseless case. The work \cite{you2016oracle} proposes to use the elastic net regularizer (mixture of two norms) to address the scalability issue.

In this paper, we present two first-order methods that can efficiently solve SSC-$\ell_1$ and SSC-$\ell_0$ optimization problems for the more general case of affine subspaces. Specifically, motivated by theoretical guarantees and empirical success of SSC-$\ell_1$, an efficient proximal gradient method is proposed  that requires $\order(n^2)$ time and $\order(n^2)$ memory 
to find the representation matrix $\CC$ for a fixed $p<n$. Another noticeable advantage of the introduced method over ADMM is the lack of additional parameter tuning for a given $\lambda_e$. 
In the case of SSC-$\ell_0$,
the main advantage of our proposed solver, 
compared to other sparse approximation techniques such as OMP, is the ability to handle the more general case of affine subspaces. 
Recent work \cite{you2019affine} showed that the affine constraint might be discarded when the ambient dimension $p$ is  large enough compared to the sum of subspace dimensions. However, this assumption is not realistic in many large-scale problems consisting of multiple subspaces. Our proposed solvers perform SSC efficiently on large data sets regardless of their ambient dimensions.

There are two prior works on using proximal gradient methods in the context of subspace clustering. For example, the authors in \cite{jiang2018nonconvex} used a proximal gradient method for the low-rank subspace clustering problem, where  nuclear norm regularization enforces the coefficient matrix $\mathbf{C}$ to be low-rank instead of being sparse. Although using the nuclear norm simplifies the problem, such  a regularizer may lead to performance degradation \cite{vidal2016sparse}, and theoretical guarantees are very limited. Another work used a proximal gradient method to solve the SSC-$\ell_0$ problem without the affine constraint \cite{yang2016ell}. Therefore, this work advances the previous research in this direction by presenting efficient solvers for the sparse subspace clustering problem even with the more general case of affine subspaces.

Additionally, we present an efficient implementation of ADMM for SSC-$\ell_1$ using the matrix-inversion lemma. The improved implementation in Remark \ref{rmk:fastADMM} reduces  the computational cost of ADMM for SSC-$\ell_1$ \cite{elhamifar2013sparse} from $\order(n^3)$ down to $\order(n^2)$. Such observations have been made for ADMM in general before, but not for SSC in particular, and many popular codes for SSC via ADMM do not use the efficient implementation.
A summary of complexity of our proposed solvers is presented in Table \ref{table:complexity}. 
{
	\begin{table*}[t]
		\caption{Summary of complexity of algorithms discussed, showing the leading order terms assuming $p<n$ where
			$\X\in\R^{p\times n}$ and \textnormal{nnz} is the number of non-zero entries in $\X$, and $k$ is the sparsity in \eqref{eq:SSCl0}. 
		}
		\newcommand{\iteration}{T}
		\renewcommand{\O}[1]{#1}
		\centering
		\begin{tabular}{lllll}
			\toprule
			\multicolumn{3}{l}{\small [$\iteration = $ \# iterations]}  & computation & memory \\
			\midrule
			& \multirow{3}{*}{linear} & ADMM~\cite{elhamifar2013sparse} & $\O{n^3+\iteration n^3}$ & $\O{n^2}$ \\   
			& & ADMM (Remark \ref{rmk:fastADMM}) & $\O{pn^2+\iteration pn^2}$ & $\O{n^2}$ \\
			SSC-$\ell_1$  & & Proposed & $\O{\iteration pn^2 }$ & $\O{n^2}$ \\
			\cmidrule{2-5}
			(Eq.\ \eqref{eq:SSC_l1})  &  \multirow{3}{*}{affine} & ADMM~\cite{elhamifar2013sparse} & $\O{n^3+\iteration n^3}$ & $\O{n^2}$ \\ 
			& & ADMM (Remark \ref{rmk:fastADMM}) & $\O{pn^2+\iteration pn^2}$ & $\O{n^2}$ \\
			& & Proposed & $\O{\iteration (p+\log n)n^2 }$ & $\O{n^2}$ \\
			\cmidrule{1-5}
			& \multirow{2}{*}{linear} & OMP~\cite{dyer2013greedy} & $\O{k(\text{nnz}\cdot n+p\sparsity n) }$ & nnz $+\O{\sparsity n}$ \\   
			SSC-$\ell_0$ & & Proposed & $\O{\iteration(\text{nnz}\cdot n +\sparsity n  ) }$ & nnz $+\O{\sparsity n}$ \\
			\cmidrule{2-5}
			(Eq.\ \eqref{eq:SSCl0}) &  \multirow{1}{*}{affine} 
			& Proposed & $\O{\iteration(\text{nnz}\cdot n +\sparsity n  ) }$ & nnz $+\O{\sparsity n}$ \\
			\bottomrule
		\end{tabular}
		\label{table:complexity}
	\end{table*}
}

The rest of the paper is organized as follows. In Section \ref{sec:review}, we provide a brief review of two main existing solvers for SSC: ADMM and OMP. Section \ref{sec:proposed} introduces the proposed proximal gradient framework along with detailed instructions on finding  proximal operators for the case of affine subspaces. In Section \ref{sec:num-exp}, we present various numerical experiments to compare our methods with the existing solvers in terms of computational savings, robustness to solver parameters, and superior performance. Concluding remarks and future research directions are given in Section \ref{sec:conc}.

\paragraph{Notation}
Lower-case and upper-case bold letters represent column vectors and matrices, respectively. For a vector $\c\in\R^n$ and $q\geq 1$, let $\|\c\|_q=(\sum_{i=1}^{n}|[\c]_i|^q)^{1/q}$ denote the $\ell_q$ norm, where $[\c]_i$ is the $i$-th element of $\c$. Also, $\|\c\|_0$ represents the $\ell_0$ pseudo-norm which counts the number of non-zero entries in $\c$. Let $\|\CC\|=\max_{\x:\|\x\|_2=1} \x^T\CC\x$ stand for the spectral norm and let $\|\CC\|_F=\sqrt{\sum_{i,j}[\CC]_{ij}^2}$ represent the Frobenius norm  with the $(i,j)$-th entry denoted by $[\CC]_{ij}$. We use the  standard matrix norm $\|\CC\|_1=\sum_{ij}|[\CC]_{ij}|$. Finally, $\diag(\CC)$ returns a column vector of the main diagonal elements of $\CC$ and $\ones$ denotes the all-ones vector of matching dimensions.

\section{Review of Sparse Subspace Clustering}\label{sec:review}
The SSC-$\ell_1$ optimization problem can be solved using generic convex solvers such as interior point methods (IPM). However, even an IPM that is customized to take advantage of problem structure would still require $\mathcal{O}(n^3)$ flops per iteration, and generally 15 to 30 iterations.
To reduce the computational cost, Elhamifar and Vidal \cite{elhamifar2013sparse} proposed to use the alternating direction method of multipliers (ADMM). Here, we briefly explain the procedure to solve SSC-$\ell_1$  via ADMM to compare with our proposed method in the next section. In our experiments, we also compare our solvers with a variant of ADMM known as Adaptive ADMM (AADMM) \cite{xu2016adaptive}, which adaptively tunes a penalty parameter to achieve fast convergence.

Let us first introduce an auxiliary matrix $\mathbf{A}\in\R^{n\times n}$ and consider the following program whose solution coincides with the solution of the original program:
\begin{align}
&\min_{\CC,\mathbf{A}}\|\CC\|_1+\frac{\lambda_e}{2}\|\X-\X\mathbf{A}\|_F^2\nonumber\\
&\text{s.t.}\; \mathbf{A}^T\ones=\ones,\;\mathbf{A}=\CC-\diag(\CC).\label{eq:ADMM1}
\end{align}
With an abuse of notation in this discussion, $\diag(\CC)$ also denotes the matrix formed by zeroing all but the diagonal entries of $\CC$.
Next, the augmented Lagrangian with $\rho>0$ is formed,
\begin{align}
\mathcal{L}(\CC,\mathbf{A},\boldsymbol{\delta},\boldsymbol{\Delta}) &= 
\|\CC\|_1+\frac{\lambda_e}{2}\|\X-\X\mathbf{A}\|_F^2+\frac{\rho}{2}h(\CC,\mathbf{A}) \ldots \notag \\
& \ldots + \boldsymbol{\delta}^T\left(\mathbf{A}^T\ones - \ones\right) + \text{trace}\left(\boldsymbol{\Delta}^T \left(\CC-\diag(\CC)\right)\right) \label{eq:ADMM-rho} \\
h(\CC,\mathbf{A}) &\defeq\|\mathbf{A}^T\ones - \ones \|_2^2+\|\mathbf{A} - (\CC-\diag(\CC))\|_F^2 \notag 
\end{align}
where the Lagrange multipliers are $\boldsymbol{\delta}\in\R^n$ and a matrix $\boldsymbol{\Delta}\in\R^{n\times n}$. 

In the $i$-th iteration of ADMM, the two matrices $\mathbf{A}$ and $\CC$ are updated sequentially (\`a la Gauss-Seidel)  by minimizing the Lagrangian with respect to the primal variables. 
Specifically, $\mathbf{A}^{(i+1)} = \argmin_{\mathbf{A}} \mathcal{L}(\CC^{(i)},\mathbf{A},\boldsymbol{\delta}^{(i)},\boldsymbol{\Delta}^{(i)})$ which can be found by solving the normal equations
\begin{align}
(\lambda_e\X^T\X+\rho\eye+\rho\ones \ones^T)\mathbf{A}^{(i+1)}=&\lambda_e\X^T\X+\rho(\ones\ones^T+\CC^{(i)})\ldots\nonumber\\& \ldots -\ones  \boldsymbol{\delta}^{(i)T}-\boldsymbol{\Delta}^{(i)},\label{eq:ADMM-update}
\end{align}
and  $\mathbf{C}^{(i+1)} = \argmin_{\mathbf{C}} \mathcal{L}(\CC,\mathbf{A}^{(i+1)},\boldsymbol{\delta}^{(i)},\boldsymbol{\Delta}^{(i)})$ which can be solved as 
$\CC^{(i+1)}=\mathbf{J} - \diag(\mathbf{J})$, where 
$\mathbf{J} = \proxx{\rho^{-1}}\left( \mathbf{A}^{(i+1)} + \rho^{-1}\boldsymbol{\Delta}^{(i)}\right)$
and 
$\proxx{\eta}$
applies to each element of the matrix and is defined as 
$\proxx{\eta}(v)=\text{sign}(v)\cdot\lfloor|v|-\eta \rfloor_+$,
with $\lfloor\tau \rfloor_+\defeq\max\{0,\tau\}$, cf.\ Eq.\ \eqref{eq:proxl1}. At the same iteration, $\boldsymbol{\delta}$ and $\boldsymbol{\Delta}$ are updated 
by a gradient descent step on the dual function:
$\boldsymbol{\Delta}^{(i+1)} = \boldsymbol{\Delta}^{(i)} + \rho\left( \mathbf{A}^{(i+1)} - \mathbf{C}^{(i+1)}\right)$ and $\boldsymbol{\delta}^{(i+1)} = \boldsymbol{\delta}^{(i)} + \rho\left( {\mathbf{A}^{(i+1)T}}\ones-\ones \right)$. 

The ADMM solver for SSC-$\ell_1$ incurs complexity $\order(n^3+n^2p)$ to form $\X^T\X$ and compute the matrix inversion for updating $\mathbf{A}$ in Eq.~\eqref{eq:ADMM-update}. If it is possible to store the resulting $n\times n$ matrix, one can apply that to the right-hand side of Eq.~\eqref{eq:ADMM-update}, which incurs complexity $\order(n^3)$ per iteration. Since the overall complexity of ADMM scales cubically with the number of data points $n$, finding subspace-preserving coefficients based on $\ell_1$ norm regularization is computationally prohibitive for large data sets. Hence, there is a need for SSC-$\ell_1$ solvers that are computationally efficient. 
\begin{remark} \label{rmk:fastADMM}
	The implementation of ADMM in \cite{elhamifar2013sparse} has $\order(n^3)$ up-front complexity cost and also $\order(n^3)$ complexity per iteration\footnote{\url{http://vision.jhu.edu/code/}} (for both linear and affine subspace clustering). However,
	by using the matrix inversion lemma (aka Sherman-Morrison-Woodbury identity), one can reduce the up-front cost to $\order(pn^2 + p^3)$ and the per-iteration cost to $\order(pn^2)$. Our numerical experiments use code from \cite{elhamifar2013sparse} with this modification. 
	Specifically, consider a simplified version of \eqref{eq:ADMM-update} as $(\X^T\X + \rho \eye ) \mathbf{A}^{(i+1)} = \widetilde{ \CC }$ where $\widetilde{ \CC }$ represents the right-hand side of  \eqref{eq:ADMM-update} and $\X$ has absorbed $\sqrt{\lambda_e}$ and appended the row $\sqrt{\rho}\ones^T$ (to account for $\rho \ones\ones^T$).
	To initialize, compute $\mathbf{M}=( \eye_{p+1} + \rho^{-1}\X\X^T)^{-1}$ (directly or implicitly via a Cholesky factorization) which costs $\order(p^2n)$ for $\X\X^T$ and $\order(p^3)$ for the inversion/factorization, then use the matrix inversion lemma 
	\begin{displaymath}
	(\X^T\X + \rho \eye )^{-1} = \rho^{-1}\eye - \rho^{-2}\X^T \mathbf{M} \X, 
	\end{displaymath}
	and never explicitly form this matrix but rather apply it to $\widetilde{ \CC }$ in $\order(pn^2+p^2n)$ time to get
	\begin{displaymath}
	\mathbf{A}^{(i+1)} = \rho^{-1}\widetilde{ \CC } - \rho^{-2}\X^T (\mathbf{M} (\X 
	\widetilde{ \CC })).
	\end{displaymath}
\end{remark}

A further disadvantage of ADMM is that 
tuning the parameter $\rho$ that was introduced in Eq.~\eqref{eq:ADMM-rho} substantially increases the computational complexity of the ADMM solver. In the implementation of SSC-$\ell_1$ solver, the regularization parameter $\lambda_e$ and the parameter $\rho$ for ADMM are controlled by a parameter $\alpha$ \cite[Prop.~1]{elhamifar2013sparse}, where $\lambda_e=\alpha/\mu$ for some $\alpha>1$, $\rho=\alpha$, and 
\begin{equation}\label{eq:mu}
\mu\defeq \min_i\max_{j\neq i} |\x_i^T\x_j|
\end{equation}
depends on the data set. In Section \ref{sec:num-exp}, we show that the choice of $\rho$ can greatly impact the performance of SSC, and that $\rho=\alpha$ is not a good choice for some data sets. Furthermore, adaptive techniques for updating the parameter $\rho$ do not address this issue.

An alternative method to reduce the memory and computational costs of SSC-$\ell_1$ is based on using $\ell_0$ norm regularization on the columns of the coefficient matrix $\CC$ \cite{dyer2013greedy}. 
Let $\sparsity$ be a pre-defined parameter that is proportional to the intrinsic dimensions of subspaces; in practice, it is a parameter that must be estimated.
For each point $\x_j$ in the data set, a $\sparsity$-sparse coefficient vector $\c_j\in\R^n$ is obtained by solving the  non-convex optimization problem in Eq.~\eqref{eq:SSCl0}.
Without the linear equality constraint for affine subspaces, the orthogonal matching pursuit (OMP) algorithm can be used to approximately solve this problem. To do so, the $j$-th column of the data matrix $\X=[\x_1,\ldots,\x_n]$ should be removed and one column of the reduced matrix is selected at a time until $k$ columns are chosen. A simple implementation of OMP requires storing $\X$ and $\order(\sparsity n)$ additional storage (for the nonzero entries of $\CC$), and incurs complexity
$\order(\text{nnz}\cdot \sparsity+\sparsity^2p)$ 
per column $j$, where $\text{nnz}\le np$ is the number of non-zero entries in $\X$.
Thus, the overall complexity of solving SSC-$\ell_0$ via OMP is a quadratic function of $n$ when the sparsity parameter $k$ is small enough compared to $n$. 

\section{The Proposed Methods}\label{sec:proposed}

Section \ref{sec:31} reviews the generic proximal gradient descent framework, and then in  \S\ref{sec:32} we show how the SSC-$\ell_1$ and SSC-$\ell_0$ problems can be solved within that framework. The methods consist of a gradient step, which is straightforward and the same for all the variants, and a proximal step. The nature of the proximal step depends on which variant of the problem we solve, and details on all four variants are in \S\ref{sec:prox}. Because we are able to fit the problems in an existing framework, we can apply standard convergence results, as discussed in \S\ref{app:A}.

\subsection{Proximal Gradient Descent Framework} \label{sec:31}
\newcommand{\ytz}{\overline{\y}}
\newcommand{\yt}{\y^{t}}
\newcommand{\ytt}{\y^{t+1}}
Our methods to solve SSC-$\ell_1$ and SSC-$\ell_0$ derive from the proximal gradient framework, which we briefly explain. For background on the convex proximal gradient algorithm see \cite{CombettesWajs05} or the book~\cite{BeckBook2017}; for background on the non-convex version, see \cite{attouch2011convergence}. The generic framework is: 
\begin{equation}
\min_{\y} \, f(\y) + g(\y)
\end{equation}
where $f$ and $g$ are both proper and lower semi-continuous (lsc) extended valued functions, and $f$ has full domain and a Lipschitz continuous gradient with Lipschitz constant $L$, and $\y$ is in a finite-dimensional Euclidean space. The function $g$ can be an indicator function $\delta_\mathcal{Y}$ of a closed non-empty set $\mathcal{Y}$ meaning that $g(\y)=0$ if $\y\in\mathcal{Y}$ and $+\infty$ otherwise.

Taking $g\equiv 0$ for the moment, observe that the basic gradient descent iteration $\ytt = \yt - \frac{1}{L}\nabla f(\yt)$ can be equivalently written as:
\[
\ytt = \argmin_{\y} \, \underbrace{f(\yt) + \nabla f(\yt)^T(\y-\yt) + \frac{L}{2}\|\y-\yt\|_2^2}_{Q_f(\y;\yt)}
\]
and that due to the smoothness assumption on $f$, $Q_f(\y;\yt) \ge f(\y) \,\forall \y$ (cf., e.g.,~\cite{NesterovBook}), so gradient descent can be viewed as minimizing a majorizing function.

Now allowing a general $g$, it immediately follows that $Q_f(\y;\yt) + g(\y) \ge f(\y)+g(\y),\forall \y$, and this motivates the update:
\begin{equation} \label{eq:update0}
\ytt \in \argmin_{\y}\, Q_f(\y;\yt) + g(\y).
\end{equation}
For any $\gamma>0$, define the proximity operator (or ``prox'' for short) to be:
\[
\prox_{\gamma g}(\ytz) \in \argmin_{\y}\, \gamma \cdot g(\y) + \frac{1}{2}\| \y - \ytz \|_2^2
\]
The minimizer may not be unique if $g$ is not convex, in which case the prox is defined as any minimizer. 
The prox is a natural extension of the Euclidean projection onto a closed nonempty set $\mathcal{Y}$, and indeed if $g$ is the indicator function of $\mathcal{Y}$ then the proximity operator is just the projection onto $\mathcal{Y}$.

By completing the square, the update \eqref{eq:update0} can be cast as: 
\begin{equation} \label{eq:update1}
\ytt = \prox_{L^{-1} g}(\yt - L^{-1}\nabla f(\yt) )
\end{equation}
which defines the generic proximal gradient algorithm.

\subsection{Algorithms for SSC-$\ell_1$ and SSC-$\ell_0$}\label{sec:32}
The proximal gradient framework applies to SSC-$\ell_1$ by identifying $f$ as:
$f(\CC) = \frac{\lambda_e}{2}\|\X-\X\CC\|_F^2$, $\mathcal{Y}_0 = \{ \CC \mid \diag(\CC)=0 \}$, $\mathcal{Y}_1 = \{ \CC \mid \CC^T\ones=\ones \}$, and 
\begin{equation}\label{eq:g1}
g(\CC) = \|\CC\|_1 + \delta_{\mathcal{Y}_0}(\CC)+ \delta_{\mathcal{Y}_1}(\CC)= \sum_{j=1}^n g_j( \c_j).
\end{equation} 
Both $f$ and $g$ are separable in the columns $\c_j$ of $\CC$ in the sense that $g(\CC) = \sum_{j=1}^n g_j( \c_j) $, and likewise for $f$.

Likewise, the framework applies to SSC-$\ell_0$ using the same $f$, and modifying $g$ to be: 
\begin{equation}\label{eq:g2}
g(\CC) = \delta_{\mathcal{Y}_{\sparsity}}(\CC) + \delta_{\mathcal{Y}_0}(\CC)+ \delta_{\mathcal{Y}_1}(\CC)= \sum_{j=1}^n g_j( \c_j)
\end{equation}
where $\mathcal{Y}_{\sparsity} = \{ \CC \mid \CC=[\c_1,\ldots,\c_n], \, \|\c_j\|_0 \le \sparsity \,\forall j=1,\ldots,n \}$. This $g$ is still separable in the columns of $\CC$.

The generic proximal gradient algorithm to solve both problems is presented in Algorithm \ref{alg:0}. We present a few standard convergence results about the proximal gradient descent algorithm in Section \ref{app:A}.

\newcommand{\Ct}{\CC^t}
\newcommand{\Ctt}{\CC^{t+1}}
\newcommand{\Ctz}{\CC^0}
\begin{algorithm}
	\caption{Prox.\ Gradient Descent for SSC-$\ell_1$ and SSC-$\ell_0$}
	\label{alg:0}
	\algrenewcommand\algorithmicensure{\textbf{Parameter:}}
	\begin{algorithmic}[1]
		\Ensure $\epsilon$ \Comment{Stopping tolerance}
		\Ensure $\Ctz$ \Comment{Initialization}
		\Require $L = \lambda_e\| \X \|^2 $ \Comment{Lipschitz constant of gradient}
		\State $t\gets 0$ \Comment{Iteration counter}
		\State $\gamma \gets L^{-1}$ (convex) or $ .99L^{-1}$ (non-convex) \Comment{Stepsize}
		\Repeat 
		\State $\widetilde{\CC} \gets \Ct - \gamma\lambda_e\X^T( \X\Ct - \X )$ \Comment{Gradient step on $f$}
		\For{ $j=1,\ldots,n$ }
		\State $\c_j^{t+1} \gets \prox_{\gamma g_j}(\tilde{\c}_j)$ \label{step:prox}
		\Comment{$g$ as in \eqref{eq:g1} or \eqref{eq:g2}}
		\EndFor
		\State $t\leftarrow t+1$
		\Until{ $\| \Ct - \Ctt \|_F \le \epsilon $ } 
	\end{algorithmic}
\end{algorithm}

\subsection{Proximity Operators for Each Case} \label{sec:prox}
We consider the computation of line~\ref{step:prox} in Algorithm \ref{alg:0} in detail, for four cases of the $g$ operator that arise from: (1) SSC-$\ell_1$ without the $\c_j^T \ones = 1$ constraint; (2)
SSC-$\ell_1$ with the $\c_j^T \ones = 1$ constraint; 
(3) SSC-$\ell_0$ without the $\c_j^T \ones = 1$ constraint; (4)
SSC-$\ell_0$ with the $\c_j^T \ones = 1$ constraint.
\begin{remark}
	All projections involve the constraint $\mathcal{Y}_0 = \{ \CC \mid \diag(\CC)=0 \}$. For a given column $\c_j$, this can be enforced by setting the appropriate entry $[\c_j]_j=0$, and working with the $n-1$ dimensional versions of the other constraints on the remaining indices. Hence, the dimensions of the columns are really $n-1$. In this section, for simplicity of exposition, we assume each column $\c_j$ has already had the appropriate entry removed, and we denote its size with $n$ rather than $n-1$.
\end{remark}
\begin{remark}
	\newcommand{\C}{\mathbf{C}}
	In all four cases for $g$, we can separate $g(\CC) = \sum_{j=1}^n g_j( \c_j) $ over the columns. The proximity operator can be computed for each $g_j$ separately and then combined (cf.\ \cite[Prop.\ 24.11]{CombettesBook2}), hence we only discuss the proximity operator for a single column $\c_j$, and denote this by $\c$ rather than $\c_j$ to unclutter notation.
	Specifically, with $\C = [\c_1,\ldots,\c_n]$, then 
	$\prox_{\gamma g}(\C) = [ \prox_{\gamma g_1}(\c_1), \ldots, \prox_{\gamma g_n}(\c_n) ]$.
\end{remark}
\subsubsection{$\ell_1$ proximity operator}
First, consider the SSC problem assuming all subspaces are true subspaces, and therefore pass through $\mathbf{0}$. In this case, there is no $\c^T \ones = 1$ constraint, and the proximity operator is:
\begin{equation}\label{eq:proxl1}
\proxx{\gamma}(\d) \defeq \argmin_{\c} \frac{1}{2}\|\c-\d\|_2^2 + \gamma \|\c\|_1
\end{equation}
and it is well-known that the solution is component-wise soft-thresholding (also known as ``shrinkage''):
\begin{equation} \label{eq:proxl1soln}
[\proxx{\gamma}(\d) ]_i = \text{sign}( d_i ) \cdot \lfloor |d_i| - \gamma \rfloor_{+}
\end{equation}
where $\lfloor \tau \rfloor_+ \defeq \max \{0,\tau\}$.

\subsubsection{$\ell_1$ proximity operator with affine constraint}
Now, consider the full SSC problem with affine spaces. The proximity operator computation is to solve:
\begin{equation}\label{ex:prox}
\argmin_{\c} \frac{1}{2}\|\c-\d\|_2^2 + \gamma \|\c\|_1 \;\text{s.t.}\; \c^T\ones = 1.
\end{equation}
Eq.~\eqref{ex:prox} is a strongly convex minimization problem with a unique solution, but it is not separable, and the solution is not-obvious, yet it clearly has specific structure. Efficient algorithms for it have been proposed going back at least to the 1980s~\cite{QP1986}, and it has been rediscovered many times (e.g., \cite{Stefanov2004,KiwielContinuousKnapsack,QP2010}). In some incarnations, it is known as the ``continuous knapsack'' problem. 
It is related to other $\ell_1$ problems, such as projection onto the $\ell_1$ ball (\cite{knapsack1984}, and re-discovered and/or improved in \cite{liu2009efficient,Duchil1,UBCl1,Pardalos1990,Maculan89}) and trust-region or exact line search variants, as well as quasi-Newton variants~\cite{quasiNewtonNIPS}. Most formulations are reducible to each other, accounting for some of the duplications in the literature. The approaches fall into a few categories: reduction to low-dimensional linear or quadratic programs, fast median searches, or one-dimensional root-finding via bisection. We present below a derivation using a one-dimensional root-finding approach that has complexity $\order{(n \log n)}$. We suspect that fast median-finding ideas might enable a $\order(n)$ algorithm but do not pursue this since theoretical $\order(n)$ median-finding algorithms are in practice slower than efficient implementations of $\order( n \log n)$ sorting algorithms until $n$ is extremely large.
\begin{prop}
	The problem \eqref{ex:prox} can be solved exactly in $\order( n \log n)$ flops.
\end{prop}
By ``exact'' solution, we mean there is no optimization error, though there is possibly roundoff error due to floating point computation unless exact arithmetic is used.
As mentioned above, related results have appeared in the literature so we do not claim novelty, but the algorithms are not well known, so we give the proof below since it also explains the algorithm.
\begin{proof}
	The standard Lagrangian for \eqref{ex:prox} is $\mathcal{L}(\c;\beta) = \frac{1}{2}\|\c-\d\|_2^2 + \gamma \|\c\|_1 + \beta ( \c^T\ones - 1 )$ where the dual variable $\beta$ is a scalar. Since the problem is convex and only has equality constraints, Slater's conditions are satisfied and the following two KKT conditions are necessary and sufficient for a point $\c$ to be optimal:
	\begin{align}
	0 &\in \partial \mathcal{L}(\c;\beta), \;\text{i.e.,}\; 0 \in \c - \d + \gamma \partial \|\c\|_1 + \beta \ones \label{eq:subdiff} \\
	&\c^T\ones = 1 \label{eq:ones}
	\end{align}
	where $\partial \mathcal{L}$ is the subdifferential. 
	Observe that Fermat's rule for convex functions, namely that $y \in \argmin F(x)$ if and only if $0\in \partial F(y)$, applied to the objective in \eqref{eq:proxl1} implies that  $\c = \proxx{\gamma}(\d)$ if and only if $ 0 \in \partial\left( \frac{1}{2}\|\c-\d\|_2^2 + \gamma \|\c\|_1 \right) = 
	\c - \d + \gamma \partial \|\c\|_1$ where the equality is true since both functions have full domain. Thus the condition \eqref{eq:subdiff} is equivalent to $\c = \proxx{\gamma}(\d - \beta \ones)$.
	Substituting this into \eqref{eq:ones} gives that
	$\ones^T\proxx{\gamma}(\d - \beta \ones) = 1$ is a necessary and sufficient condition in terms of only the scalar $\beta$. We can rewrite this condition as:
	\begin{equation}\label{eq:scalar}
	0 = f(\beta) 
	\defeq \sum_{i=1}^n \text{sign}(d_i-\beta)\cdot\lfloor |d_i-\beta| -\gamma \rfloor_+ -1.
	\end{equation}
	This is a one-dimensional, piecewise linear root-finding problem in $\beta$, and the linear regions occur between the break-points where $|d_i-\beta|=\gamma$, i.e., $\beta = d_i \pm \gamma$. In the linear regions, solving for $\beta$ is just solving a 1D linear equation, so the only difficulty is  finding the correct linear region. Each term in the sum of $f$ is monotonically decreasing in $\beta$, therefore the function $f$ is monotonically decreasing in $\beta$. There are $2n$ break-points of the form $\beta = d_i \pm \gamma$, so our algorithm sorts these $2n$ break-points, with cost $\order( n \log n)$ (e.g., using merge sort), and then does a bisection search on the regions defined by the break-points, with $\order( \log n)$ steps, and linear complexity per step. See Algorithm \ref{alg:1}.
\end{proof}

\begin{algorithm}
	\caption{Algorithm to solve Eq.~\eqref{ex:prox}}
	\label{alg:1}
	\newcommand{\upper}{i_\text{max}}
	\renewcommand{\lower}{i_\text{min}}
	\begin{algorithmic}[1]
		\State $\proxx{\gamma}$ defined as prox from Eq.~\eqref{eq:proxl1soln}
		\State Convention: $b_0=-\infty, b_{2n+1}=+\infty$
		\Function{Prox}{$\d\in \R^n, \gamma\in\mathbb{R}^+$}
		\State $\lower = 0, \upper = 2n+1$
		\State $\mathbf{b}=\text{sort}( \{\d-\gamma\}\cup\{\d+\gamma\} )$
		\Comment{ $b_1 \ge b_2 \ldots \ge b_{2n}$}
		\While{ $\upper - \lower > 1$ }
		\State $j \gets \lfloor (\lower + \upper)/2 \rfloor$ \Comment{Round to an integer}
		\State $\c \gets \proxx{\gamma}( \d - b_j\ones )$
		\If{ $\c^T\ones > 1$ }
		$\upper \gets j$
		\Else\ 
		$\lower \gets j$
		\EndIf
		\EndWhile
		\State Choose any $\beta\in (b_{\lower}, b_{\upper})$
		\State $\c \gets \proxx{\gamma}( \d - \beta\ones )$
		\State $\sss^\Ast \gets \text{supp}(\c)$ \Comment{Find the support}
		\State $\beta^\Ast \gets \frac{-1}{|\sss^\Ast|}\left(
		1-\sum_{i\in\sss^\Ast} d_i - \gamma\text{sign}(c_i)  \right)$
		\State $\c \gets \proxx{\gamma}( \d - \beta^\Ast\ones )$
		\State \Return $\c$
		\EndFunction
	\end{algorithmic}
\end{algorithm}
\subsubsection{$\ell_0$ projection}
Again, we first discuss the problem assuming all subspaces are true subspaces and not affine spaces, so there is no $\c^T \ones = 1$ constraint.
The relevant proximity operator reduces to the following Euclidean projection:
\newcommand{\projj}{\proj_k}
\begin{equation}\label{eq:projl0}
\argmin_{\c} \frac{1}{2}\|\c-\d\|_2^2 \;\text{s.t.}\;\|\c\|_0 \le \sparsity.
\end{equation}
While this is a non-convex problem, due to its simple structure, it is easy to solve. For example, one can sort the absolute value of all $n$ terms $(|d_i|)$ and then choose the top $\sparsity$ largest (which may not be unique if there are duplicate values of $|d_i|$), at cost $\order(n\log(n))$. Alternatively, it may be faster to take the  largest entry in absolute value, and repeat $\sparsity$ times $\order( n\sparsity )$. Specialized implementations based on heapsort can also return the answer in $\order( n\log(\sparsity) )$~\cite{knuth1997art}.

\subsubsection{$\ell_0$ projection with affine constraint}
Adding in the affine constraint $\c^T \ones = 1$, the relevant proximity operator is:
\begin{equation}\label{eq:projl0_2}
\argmin_{\c} \frac{1}{2}\|\c-\d\|_2^2 \;\text{s.t.}\;\|\c\|_0 \le \sparsity, \; \c^T \ones = 1.
\end{equation}
\newcommand{\observation}{\d}
\newcommand{\obs}{d}
It is not obvious that there is an efficient algorithm to solve this non-convex problem, but in fact due to its special structure, there is a specific greedy algorithm, known as the ``greedy selector and hyperplane projector'' (GSHP), which has been shown to exactly solve \eqref{eq:projl0_2} and take time complexity $\order( n \cdot k)$~\cite{sparseSimplexICML}; pseudo-code is shown in Algorithm \ref{alg:2}.
For a set $\sss$ and vector $\observation\in\R^n$, the notation $\restr{\observation}{\sss}$ refers to the vector created by restricting $\observation$ to the entries in $\sss$, and $\sss^c = \{1,2,\ldots,n\}\setminus \sss$.

\begin{algorithm}
	\newcommand{\signal}{\c}
	\caption{GSHP to solve Eq.~\eqref{eq:projl0_2}~\cite{sparseSimplexICML}}
	\label{alg:2}
	\begin{algorithmic}[1]
		\State $\proj(\observation) \defeq 
		\observation - \frac{1}{n}(\observation^T\ones - 1)\ones $ \ \Comment{Proj.\ onto $\{\signal\mid \signal^T\ones=1\}$}
		\Function{GSHP}{$\observation\in \R^n, k\in\mathbb{N}^+$}
		\State $\ell=1$ , $\sss = j, \quad j \in \arg\max_i \left[ \obs_i\right]$ \Comment{Initialize}
		\Repeat \ 
		$\ell \leftarrow \ell+1, \sss\leftarrow \sss \cup \{j\}$, where \\ $\qquad j\in \arg\max_{i\in \sss^c} \left|\obs_i - \frac{\sum_{j \in \sss}\obs_j-1}{\ell-1}\right|$ \Comment{Grow}
		\Until{ $\ell= \sparsity$, set $\sss^\Ast\leftarrow \sss$  }
		\State $\restr{\signal}{\sss^\Ast} = \proj( \restr{\observation}{\sss^\Ast}),
		\;  \restr{\signal}{(\sss^{\Ast})^c} = 0 $ \Comment{Final projection}
		\State \Return $\signal$
		\EndFunction
	\end{algorithmic}
\end{algorithm}

\subsection{Convergence Results} \label{app:A}
In this section, we provide convergence results for the proposed SSC-$\ell_1$ and SSC-$\ell_0$ solvers.

\subsubsection{SSC-$\ell_1$}
\begin{thm}
	Let $(\Ct)_{t\in\mathbb{N}}$ be the sequence of points generated by Algorithm \ref{alg:0},  let $\CC^\star$ be any optimal solution to SSC-$\ell_1$ \eqref{eq:SSC_l1},
	and let $F(\cdot)$ denote the objective function in \eqref{eq:SSC_l1}. Then for any $t\in \mathbb{N}$, $\Ct$ is feasible for \eqref{eq:SSC_l1} and
	\[
	F(\Ct) - F(\CC^\star) \le \frac{L}{2}\frac{1}{t}\| \Ctz - \CC^\star \|_F^2.
	\]
	Furthermore, $(\Ct)_{t\in\mathbb{N}}$ converges to an optimal point.
\end{thm}
This is a well-known result. See, for example, the textbook \cite[Thm. 10.21]{BeckBook2017} for the rate, and the textbook \cite[Cor.\ 28.9]{CombettesBook2} for the sequence convergence.
We present this result for simplicity, but note that ``Nesterov accelerated'' variants of proximal gradient descent (also known as ``FISTA'') have a very similar per-step computational cost and improve the convergence rate to $\mathcal{O}(1/t^2)$ instead of $\mathcal{O}(1/t)$. There are also variants that allow for variable step-sizes, rather than just $1/L$.
If $\gamma=1/L$ is used, $L$ is not needed to high accuracy, so it can be computed with a few iterations of the power method, or exactly in $\order(p^2n)$ time. 
In practice, for the SSC-$\ell_1$ problem, we use the Nesterov accelerated variants provided in the TFOCS package~\cite{templates}
which also incorporates a line search for the stepsize.

\begin{remark}
	Note that Algorithm \ref{alg:0} solves for all columns of $\CC$ at once, requiring $\mathcal{O}(n^2)$ memory. If memory is a concern, the problem can be solved a single column at a time due to its separable nature, requiring only $\mathcal{O}(pn)$ memory (to store $\X$) for \eqref{eq:SSC_l1} or $\order(\text{nnz}(\X)+\sparsity n)$ for \eqref{eq:SSCl0}, and not changing the asymptotic computational cost. This should not be done unless necessary, since computing with all blocks at once allows for efficient level-3 BLAS operations which are optimized to reduce communication cost and greatly improve practical performance. In practice, a few columns at a time can be solved.
\end{remark}

\begin{remark}
	The convergence results for both convex and non-convex cases do not change whether one includes the $\c_j^T \ones = 1$ constraint or not. Dropping the constraint only simplifies the computation of the proximity operator, as discussed in Section \ref{sec:prox}.
\end{remark}

\subsubsection{SSC-$\ell_0$}
This is a non-convex problem, so one would not expect \emph{a priori}  global convergence guarantees. In particular, we cannot guarantee that for an arbitrary initialization, the sequence converges to a global minimizer, but the following theorem does show that the algorithm is at least \emph{consistent} with the optimization problem. 
The theorem is actually unusually strong for non-convex problems, and relies on the results by Attouch et al. \cite{attouch2011convergence} on the Kurdyka-\L{}ojasiewicz inequality. More traditional theory would only have been able to guarantee that, at best, any cluster point of the sequence is a stationary point of the optimization problem.  
\begin{thm}\label{thm:sscl0}
	Let $(\Ct)_{t\in\mathbb{N}}$ be the sequence of points generated by Algorithm \ref{alg:0}.
	If the sequence $(\Ct)_{t\in\mathbb{N}}$ is bounded, then it converges to a stationary point $\overline{\CC}$ of SSC-$\ell_0$ \eqref{eq:SSCl0}, i.e., $\overline{\CC}$ is feasible and 
	\[
	-\nabla f( \overline{\CC} ) \in N( \overline{\CC} )
	\]
	where $N$ is the normal cone of the set $\mathcal{Y} = \mathcal{Y}_k \cap \mathcal{Y}_0 \cap \mathcal{Y}_1$, i.e.,
	\[
	\nabla f(\overline{\CC})^T(\CC - \overline{\CC}) \ge 0\;\forall \CC\in\mathcal{Y}
	\]
\end{thm}    
The proof follows from using $\epsilon=.01/L$ in \cite[Thm.\ 5.3]{attouch2011convergence} and observing that $f$ and $g$ are semi-algebraic and all the sets $\mathcal{Y}$ are closed. 

\begin{remark}
	As in the convex case, we can solve for each column $\c_j$ one-by-one. If $\X$ is sparse, the memory savings are potentially very large, since for a single column, we only need a temporary memory of $\mathcal{O}(n)$ and $\mathcal{O}( \text{nnz}(\X) + \sparsity )$ for the variables.
\end{remark}

\section{Numerical Experiments}\label{sec:num-exp}
We compare the performance of our proposed methods from Section \ref{sec:proposed} with ADMM and OMP.  Most of our experiments focus on the affine case, since there are fewer algorithms available to solve it, and some authors argue it is more powerful since it is a more general model.
We implemented the proximal operators in MATLAB and C++, and then incorporated these into the generic proximal minimization framework of the software package TFOCS \cite{templates}. We write TFOCS in the legend of figures to mean our implementation of Algorithm \ref{alg:0} for either the SSC-$\ell_1$ or SSC-$\ell_0$ case.

\begin{algorithm}
	\caption{End-to-end algorithm including spectral clustering}
	\label{alg:overall}
	\algrenewcommand\algorithmicensure{\textbf{Parameter:}}
	\begin{algorithmic}[1]
		\Ensure $K$ \Comment{Estimated number of clusters}
		\Ensure $\lambda_e$ \Comment{For SSC-$\ell_1$ only}
		\State $\CC\in\R^{n\times n}\gets \texttt{Algorithm 1}$ \Comment{SSC-$\ell_1$ or SSC-$\ell_0$, affine or not}
		\State $\W \gets |\CC| + |\CC|^T$ \Comment{Often very sparse}
		\State $[\mathbf{D}]_{ii} = \sum_{j=1}^n [\mathbf{W}]_{ij}$, $\mathbf{D}\in\R^{n\times n}$ diagonal
		\State $\mathbf{V} \gets \texttt{eig}( \mathbf{D}^{-\frac{1}{2}}\W\mathbf{D}^{-\frac{1}{2}}, K )$,
		$\mathbf{V}=[\vv_1^T; \ldots; \vv_n^T]\in\R^{n \times K}$ \Comment{Only need eigenvectors corresponding to $K$-largest eigenvalues}
		\State $\vv_i \gets \vv_i/\|\vv_i\|_2$ for $i = 1,\ldots,n$
		\State Cluster via $\texttt{kmeans}( \{\vv_i\}_{i=1}^n, K )$
	\end{algorithmic}
\end{algorithm}

The full clustering algorithm is shown in Alg.~\ref{alg:overall}, which consists of running one of the four optimization solver described in the previous section, followed by spectral clustering. When $n$ is large, we use Matlab's Krylov-subspace based solver \texttt{eigs} to compute the eigenvalue decomposition. The final K-means clustering is done via Matlab's \texttt{kmeans} which uses Lloyd's algorithm and takes the best of 20 random initializations.

As explained in Remark \ref{rmk:fastADMM}, the implementation of ADMM in \cite{elhamifar2013sparse} has $\order(n^3)$ complexity. However, we  provided a more efficient implementation using the matrix-inversion lemma that has reduced the per-iteration cost to $\order(n^2)$. The regularization parameter $\lambda_e$ for SSC-$\ell_1$  is controlled by some parameter $\alpha>1$ as $\lambda_e=\alpha/\mu$, where $\mu$ is a quantity that depends on the given data set (cf.~Eq.\ \eqref{eq:mu}). In all experiments with $\ell_1$ norm regularization, TFOCS and ADMM share the same regularization parameter $\lambda_e$. However, ADMM requires the additional parameter $\rho$ to be tuned. The default value for $\rho$ in the implementation provided by the authors is $\rho=\alpha$. 
In agreement with the findings of many other papers, we observe that
the choice of $\rho$ can greatly impact the performance of ADMM. 
Thus, one should ideally tune the parameter $\rho$ for each experiment, which increases the overall computational cost of ADMM for SSC-$\ell_1$. 
We also show that our proposed solver outperforms a new variant of ADMM, called ``Adaptive-ADMM'' (AADMM), which adaptively tunes the parameter $\rho$ for fast convergence \cite{xu2016adaptive}.

Throughout this section, we use real and synthetic and data sets. The first real data set is the Extended Yale B data set \cite{georghiades2001few}. This data set contains frontal face images of $38$ individuals under $64$ different illumination conditions. These images are downsampled to $48\times 42$ pixels, thus the data points lie in $\R^p$ with $p=2,\!016$.

The second real data set is the CoverType data set\footnote{http://archive.ics.uci.edu/ml/datasets/Covertype} which contains $n=581,\!012$ observations of $p=54$ features, where each observation is the forest cover type (lodgepole pine, cottonwood/willow, etc.) of a $30$m by $30$m section of Earth, and examples of features are elevation, aspect, etc. There are $K=7$ possible forest cover types.

The synthetic data is based on the following statistical model that considers $n$ data points in $\R^p$ drawn from a union of $K$ affine subspaces $\{\mathcal{S}_l\}_{l=1}^K$: 
\begin{equation}
\x_i=\UU^{(l)}\z_i+\MU^{(l)}+\vv_i,\;\;\forall \x_i\in\SSS_l,\label{eq:syn}
\end{equation}
where the columns of $\UU^{(l)}\in\R^{p\times r_l}$ form an orthonormal basis of $\SSS_l$, $\z_i\in\R^{r_l}$ is the low-dimensional representation of $\x_i$ with respect to $\UU^{(l)}$, $\MU^{(l)}\in\R^p$ is the intercept of $\SSS_l$, and $\vv_i\in\R^p$ is the noise vector. Thus, we can control the number of subspaces, their dimensions, intersections, and the amount of noise in order to gain insights on the performance of the aforementioned solvers. We test various SSC-$\ell_1$ solvers on up to $n=15,\!000$ data points.

\subsection{SSC-$\ell_1$ on the Extended Yale B Data Set}
In the first experiment, we compare the performance of the proposed TFOCS solver with ADMM and adaptive ADMM (AADMM) for solving SSC-$\ell_1$ on the Extended Yale B data set when the parameter $\alpha$ is set to be $1.1$. For ADMM, we consider the recommended value of $\rho$, $\rho=\alpha$, as well as the alternatives $\rho=10\alpha,100\alpha$.
Three metrics are used to demonstrate the performance of these solvers over $100$ iterations (we report all three metrics because in our experience they are not necessarily correlated with each other):  
(1) value of the objective function in Eq.~\eqref{eq:SSC_l1}; (2) subspace preserving error \cite{you2016scalable}, which is the average fraction of $\ell_1$ norm of each representation vector in the data set that comes from other subspaces; and (3) clustering error, which is the fraction of misclustered points after applying spectral clustering to $\W$ \cite{heckel2015robust}.

Since we want to compare the three solvers in each iteration and the solution of ADMM is not necessarily feasible (e.g., $\c_j^T\ones$ may not be $1$), we find the closest feasible solution by first removing the $j$-th element of $\c_j$ to get $\bar{\c}_j\in\R^{n-1}$. Then, we solve the following:
\begin{equation}
\c_j^\Ast=\argmin_{\c\in\R^{n-1}}\frac{1}{2}\|\c-\bar{\c}_j\|_2^2\;\text{s.t.}\;\c^T\ones=1.\label{eq:feas}
\end{equation}
It is straightforward to show that the solution of this problem is $\c_j^\Ast=\bar{\c}_j-\nu\ones$, where the scalar is $\nu=(\bar{\c}_j^T\ones-1)/(n-1)$. The feasible representation vectors are only used for evaluating the three metrics in each iteration and they are not used for next iterations of ADMM.

\begin{figure*}[t]
	\centering
	
	\subfloat[$K=2$]{
		\includegraphics[width=1\textwidth]{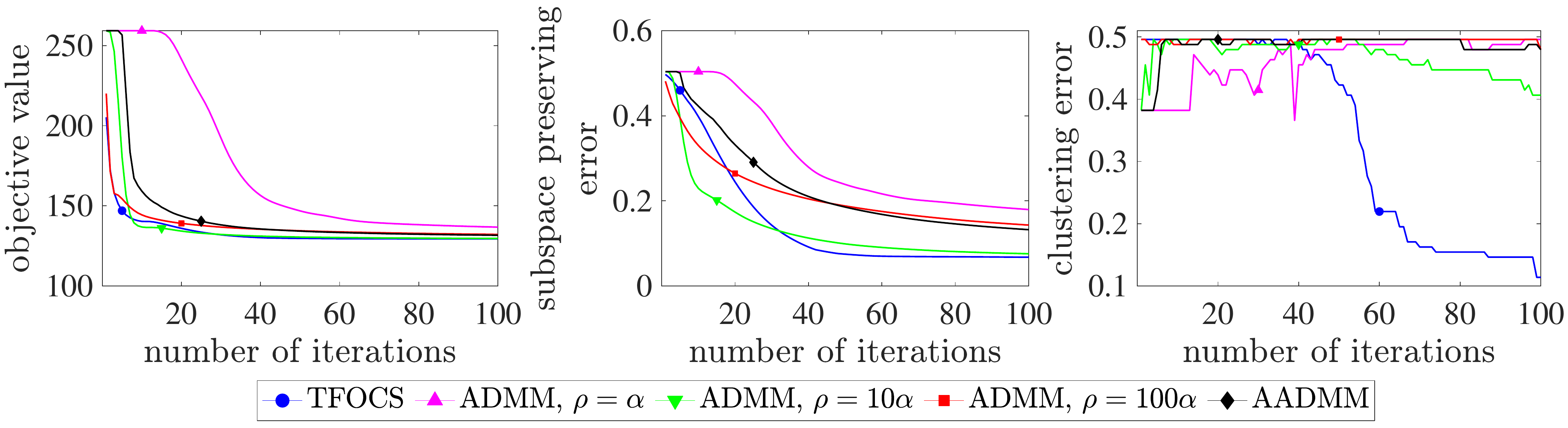}
		\label{fig:K2}
	}
	
	\subfloat[$K=3$]{
		\includegraphics[width=1\textwidth]{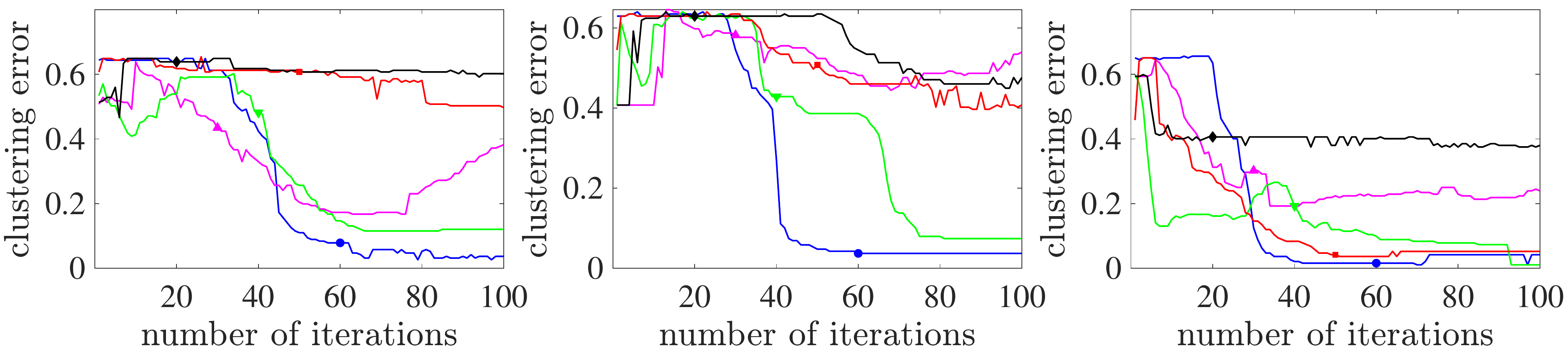}
		\label{fig:K3}
	}
	\caption{SSC-$\ell_1$ on the Extended Yale B data set for (a) $K=2$ and (b) $K=3$ clusters. For each case, three metrics are used from left to right: value of the objective function, subspace preserving error, and clustering error. The legends in (a) and (b) are the same. 
		\label{fig:yale}}
\end{figure*}

In Figure \ref{fig:K2}, the three metrics are plotted when $K=2$ clusters are selected uniformly at random from $38$ individuals. It is observed that the performance of ADMM depends heavily on the choice of the penalty parameter $\rho$. 
Interestingly, the choice of $\rho=\alpha$ is found to result in the worst performance. 
However, our proposed solver outperforms or has similar performance compared to ADMM without having to tune additional parameters. Moreover, the recently proposed AADMM which adaptively tunes $\rho$ seems to be effective, but does not compete with our proposed solver. 
We also 
report clustering errors in Figure \ref{fig:K3} for three independent trials when $K=3$ clusters are randomly selected.  Similar results are obtained when the feasibility projection in Eq.~\eqref{eq:feas} is not performed. 

We note that the clustering error we found for $K=2$ is 
higher than found in the original sparse subspace clustering (SSC) paper \cite{elhamifar2013sparse}. The reason is that like many other papers, we used a subset of $K$ individuals from the entire face data set. Thus the clustering error depends heavily on which subset is chosen (in general, clustering error depends on the orientation of subspaces). To illustrate our point, we used the original SSC code (and the values that were originally recommended) and we observed that for $K=2$, the clustering error can be as high as 0.5 depending on the selected subset. 

\subsection{Varying Values of $\rho$ in ADMM}
\begin{figure}
	\centering
	\includegraphics[width=0.5\textwidth]{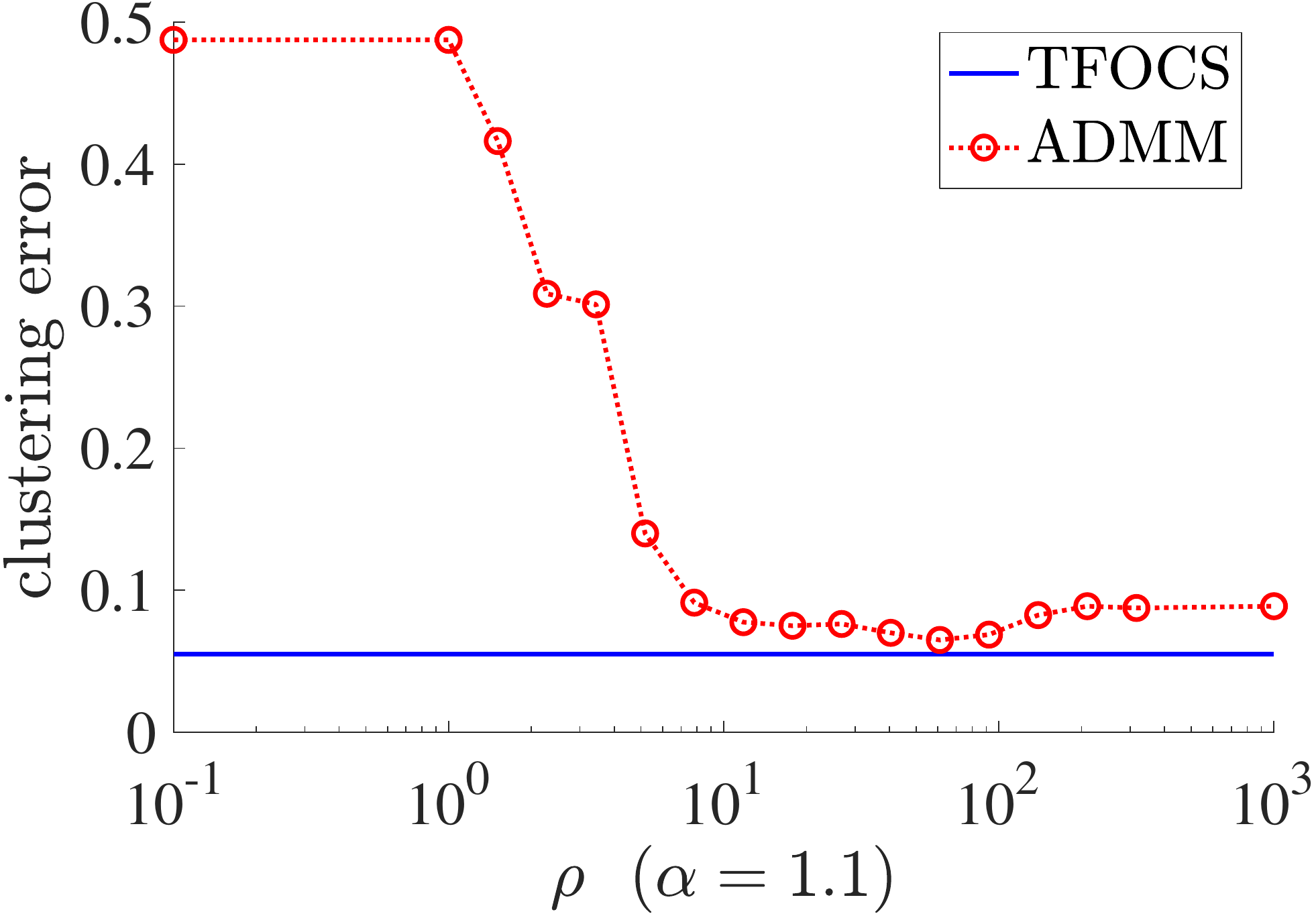}    
	\caption{Clustering error as a function of $\rho$.\label{fig:cover}}
\end{figure}

We note that larger values of $\rho$ for ADMM does not necessarily improve performance. To demonstrate this point, we compare the performance of TFOCS and ADMM solvers for SSC-$\ell_1$ when the maximum number of 
iterations is set to be $250$. 
We set $\alpha=1.1$ and consider various values of $\rho$ 
from $0.1$ to $1,\!000$ (approximately from $0.09\alpha$ to $909\alpha$) 
for a subset of $K=2$ clusters with $400$ data points chosen uniformly at random from each cluster of the CoverType data set. The clustering error results are shown in Figure \ref{fig:cover}. As we see, the performance of ADMM is close to our solver for a small interval of $\rho$, which again emphasizes the importance of tuning $\rho$ for any given data set. 

\subsection{SSC-$\ell_1$ on Synthetic Data Sets}
We consider the statistical model described in Eq.~\eqref{eq:syn}. This model allows us to control the number of subspaces $K$, their dimensions $r_l$, orientations, and the amount of noise. We set parameters $p=256$, $K=10$, $r_l=3$, and $\MU^{(l)}=\mathbf{0}$ for all $l\in\{1,\ldots,10\}$. The columns of the orthonormal matrices $\UU^{(l)}\in\R^{p\times r_l}$ are drawn uniformly at random from a set of $p$ orthonormal random vectors in $\R^p$. Each coefficient vector $\z_i\in\R^{r_l}$ is drawn i.i.d.~from the standard normal distribution. The noise vectors $\vv_i\in\R^p$, $i=1,\ldots,n$, are drawn i.i.d.~according to $\mathcal{N}(\mathbf{0},\sigma^2\eye)$, where we set $\sigma=0.1$. We sample $600$ to $1,\!500$ data points per subspace, which leads to the total number of data points from $n=6,\!000$ to $n=15,\!000$. 

\begin{figure}
	\centering
	\includegraphics[width=0.5\textwidth]{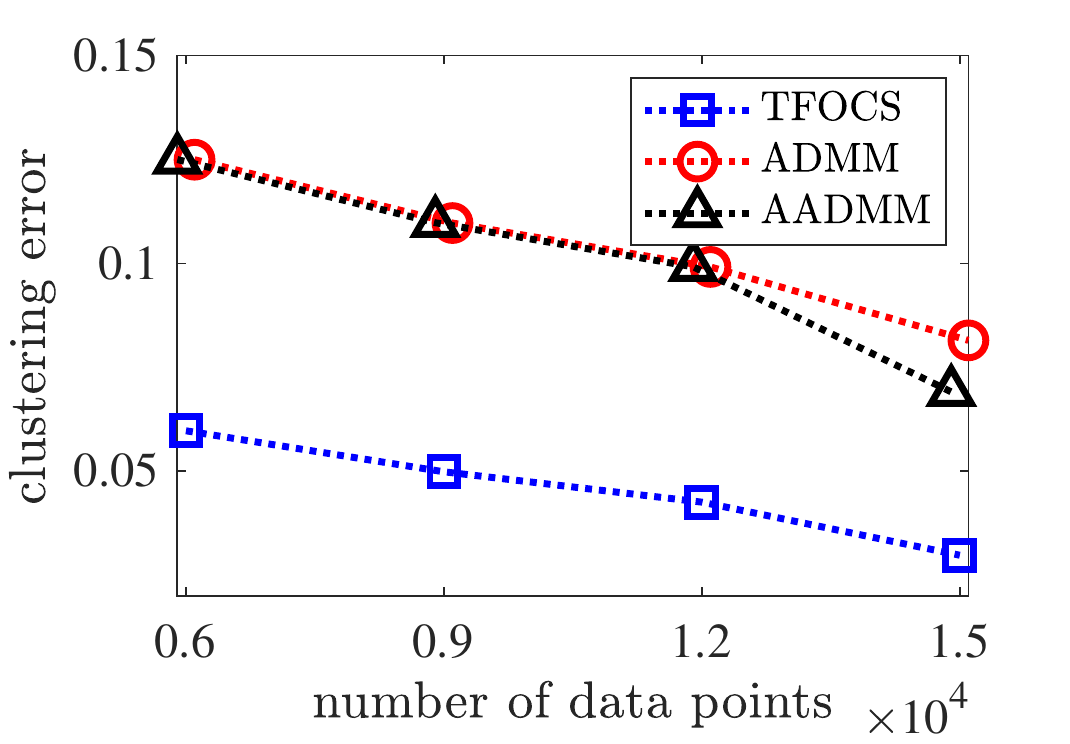}
	\caption{Clustering error of SSC-$\ell_1$ on synthetic data.  \label{fig:ce-large}}
\end{figure}

The clustering error results averaged over $10$ independent trials are presented in Figure \ref{fig:ce-large} for fixed $\alpha=30$, $\rho=10 \alpha$ for ADMM, and the maximum number of iterations is set to be $50$. We observe that our solver consistently outperforms both ADMM and AADMM. For example, when $n=15,\!000$, the average errors are $0.03$, $0.08$, and $0.07$ for our solver, ADMM, and AADMM, respectively.

\begin{figure}
	\centering
	\includegraphics[width=0.5\textwidth]{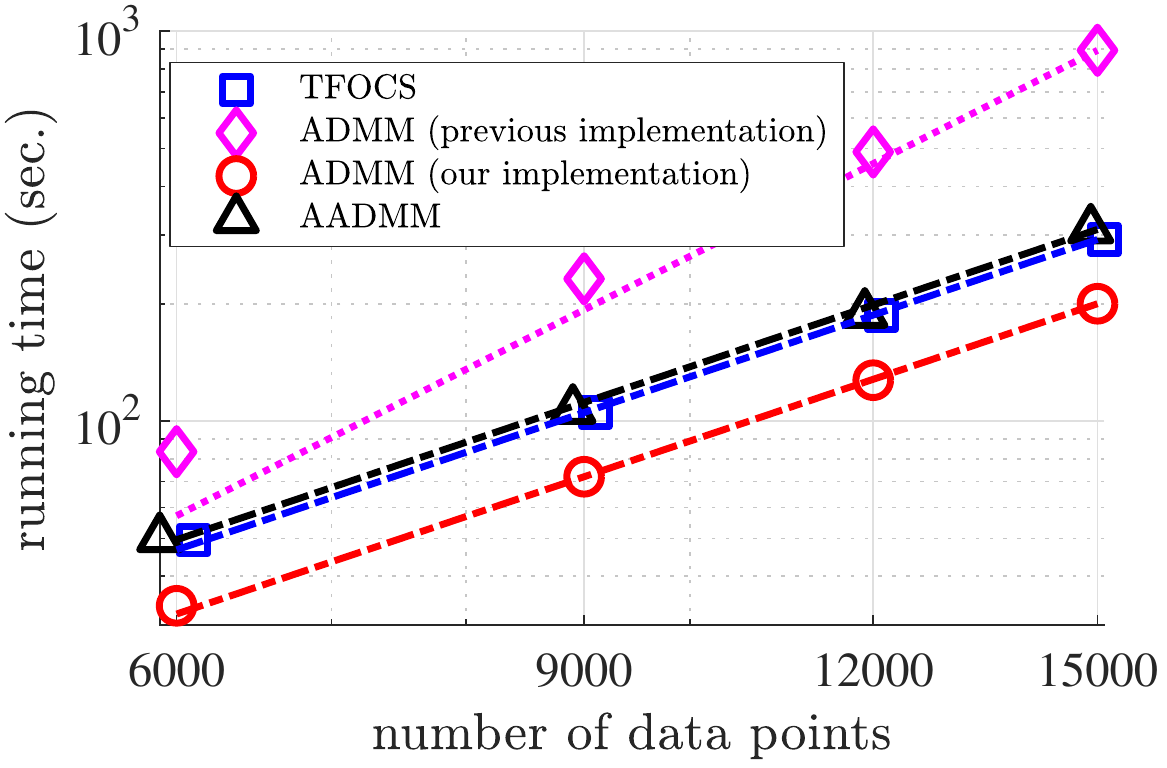}
	\caption{Running time (logarithmic scale) of SSC-$\ell_1$ on synthetic data for varying $n$.  \label{fig:time-large}}
\end{figure}

To demonstrate the efficiency of the SSC-$\ell_1$ solvers, the average running times in seconds are plotted in Figure \ref{fig:time-large}. These results verify our claim that both the proposed TFOCS solver and our implementation of ADMM scales quadratically with the number of data points $n$. However, the implementation of ADMM in \cite{elhamifar2013sparse} has complexity $\order(n^3)$.  
Although, the new implementation of ADMM is slightly faster than our proposed TFOCS solver by a constant factor, its performance depends crucially on the parameter $\rho$. Therefore, one can argue that the effective cost of ADMM is higher compared to the proposed solver in this work as our solver does not require any additional parameter tuning.

\subsection{SSC-$\ell_1$ and SSC-$\ell_0$ on the CoverType Data Set}
We test the subspace $\ell_0$ model on the CoverType data set with $n=581,\!012$, $p=54$ and  $K=7$.  Due to the size of $n$, the variable $\mathbf{C} \in \mathbb{R}^{n \times n}$ can never be formed except as a sparse matrix.
It was reported in \cite{you2016oracle} that for this data set, even using just two iterations of the solvers, that OMP took $783$ minutes, 
and two SSC-$\ell_1$ methods (one based on the original SSC ADMM algorithm, without using Remark~\ref{rmk:fastADMM}) either did not finish the two iterations within 7 days, or used  more than 16 GB of memory.

The results of running our models on this data (after normalizing the features to z-scores), and taking 2 steps as in  \cite{you2016oracle}, is presented in Table~\ref{table:2}.
We do not include results for SSC-$\ell_1$ with affine constraints, as with default parameters  this model does not lead to a sparse $\mathbf{C}$, and hence there are memory issues. The SSC-$\ell_0$ models are guaranteed to give a sparse output, and we test the non-affine variant on the full CoverType data, in addition to give results for randomly subsampling $n=10^5$ data points (about $1$ in $5$). For SSC-$\ell_1$, $\alpha$ was set to $0.1$ to encourage sparsity.

The time for the full $n=5.81\cdot10^{5}$ data is $30.4\times$ slower than for the $n=10^{5}$ data. Based on the $\mathcal{O}(n^2)$ complexity, one would expect it to be $33.7\times$ slower, which is in good agreement (to within factors such as cost of memory movement, CPU throttling and efficiencies of scale). It is notably faster than all the methods discussed in \cite{you2016oracle}. The experiment was run on a 6-core 2.6~GHz laptop with 16 GB of RAM.

\begin{table}
	\small
	\begin{tabular}{llllll}
		\toprule
		& Algorithm & Time & $\text{Time}_\text{SC}$ & Avg sparsity & Accuracy \\
		\cmidrule{2-6}
		\multirow{3}{*}{$n=10^5$}   & SSC-$\ell_1$, not affine & $3.7$ & $0.04$ & $35$ &  $42.24\%$    \\
		& SSC-$\ell_0$, not affine & $4.6$ & $0.15$ & $7$ &  $41.41\%$ \\
		& SSC-$\ell_0$, affine & $4.6$ & $0.15$ & $7$ & $43.66\%$ \\
		\cmidrule(l){2-6} 
		$n=5.81\cdot10^{5}$ &  SSC-$\ell_0$, not affine & $139.9$ & $2.7$ & $7$ & $43.41\%$ \\
		\bottomrule
	\end{tabular}
	\caption{Results on CoverType data, $p=54$. 
		Times are in minutes. $\text{Time}_\text{SC}$ is the time for the spectral clustering step. Avg sparsity is the avg number of nonzero entries per column of $\mathbf{C}$.
		There are 7 types of data, so accuracy for random guessing is $14\%$. \label{table:2}
	}
\end{table}

\subsection{SSC-$\ell_0$ on Synthetic Data Sets}
In this experiment, we again use a synthetic data set generated based on the statistical model described in Eq.~\eqref{eq:syn}. The parameters are $p=64$, $K=3$, $r_l=10$, $n=600$, and $\MU^{(l)}=\mathbf{0}$ for all $l\in\{1,2,3\}$. 
We choose $\MU^{(l)}=\mathbf{0}$, i.e., subspace not affine space clustering, since we do not know of other algorithms that can handle the affine space case.
The maximum number of iterations is set to be $100$. Similar to one of the experiments in \cite{heckel2017dimensionality}, we consider the case that every pair of subspaces intersects in at least $5$ dimensions. To do so, the orthonormal bases are given by $\UU^{(l)}=[\UU\;\; \widetilde{\UU}^{(l)}]\in\R^{p\times 10}$, where matrices $\UU$ and $\widetilde{\UU}^{(l)}$, $l=1,2,3$,  are chosen uniformly at random among all orthonormal matrices of size $p\times 5$. The noise term $\vv_i\in\R^p$ is distributed according to $\mathcal{N}(\mathbf{0},\sigma^2\eye)$, where the noise level  $\sigma$ is varied from $0$ to $1.0$. 
Therefore, this synthetic data set allows us to study the impact of noise as well as the choice of $k$ on clustering performance using our TFOCS and OMP methods for solving SSC-$\ell_0$.

\begin{figure*}
	\centering
	\subfloat[$k=10$]{
		\includegraphics[width=.49\textwidth]{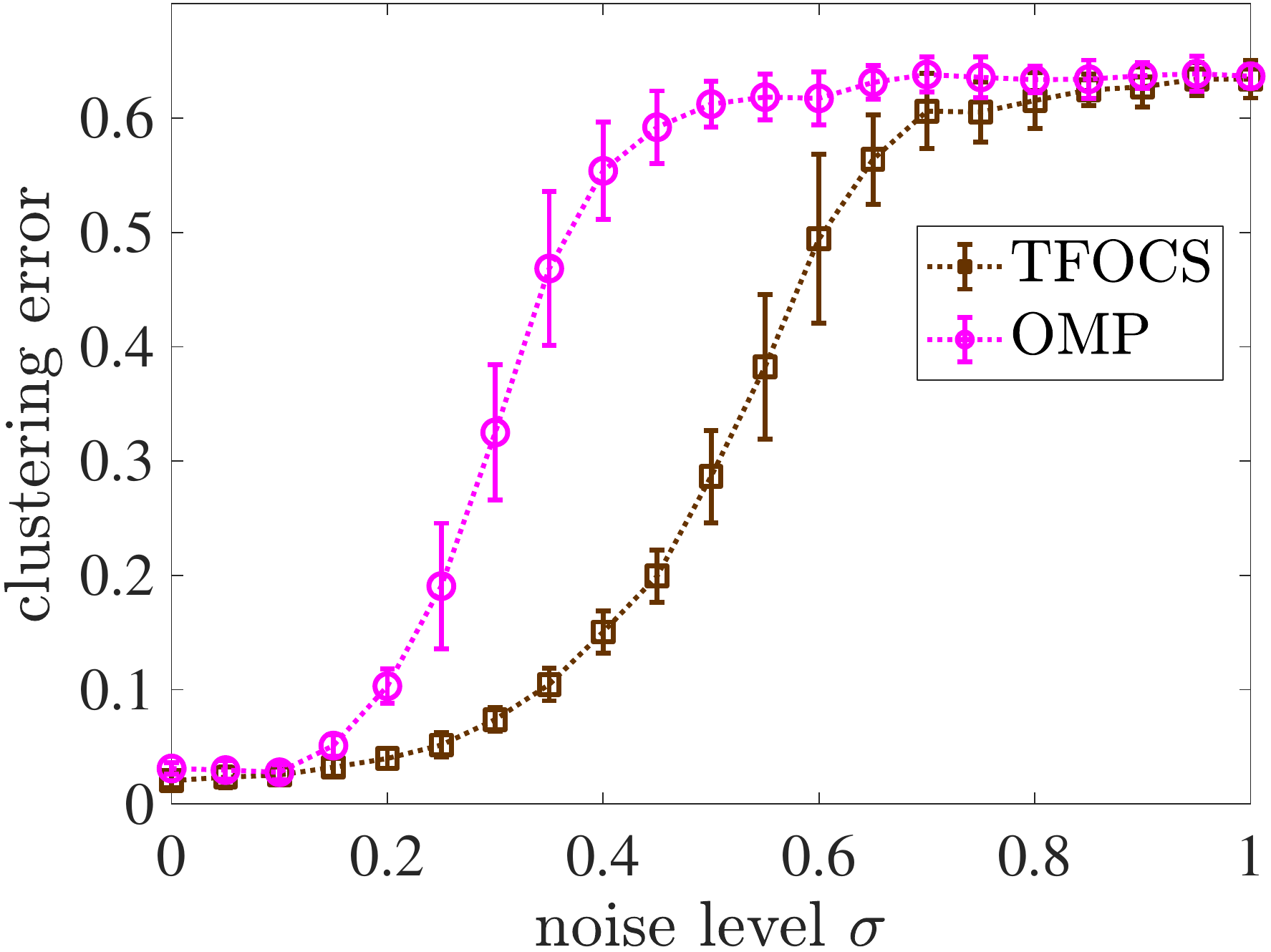}
	}
	\subfloat[$k=20$]{
		\includegraphics[width=.49\textwidth]{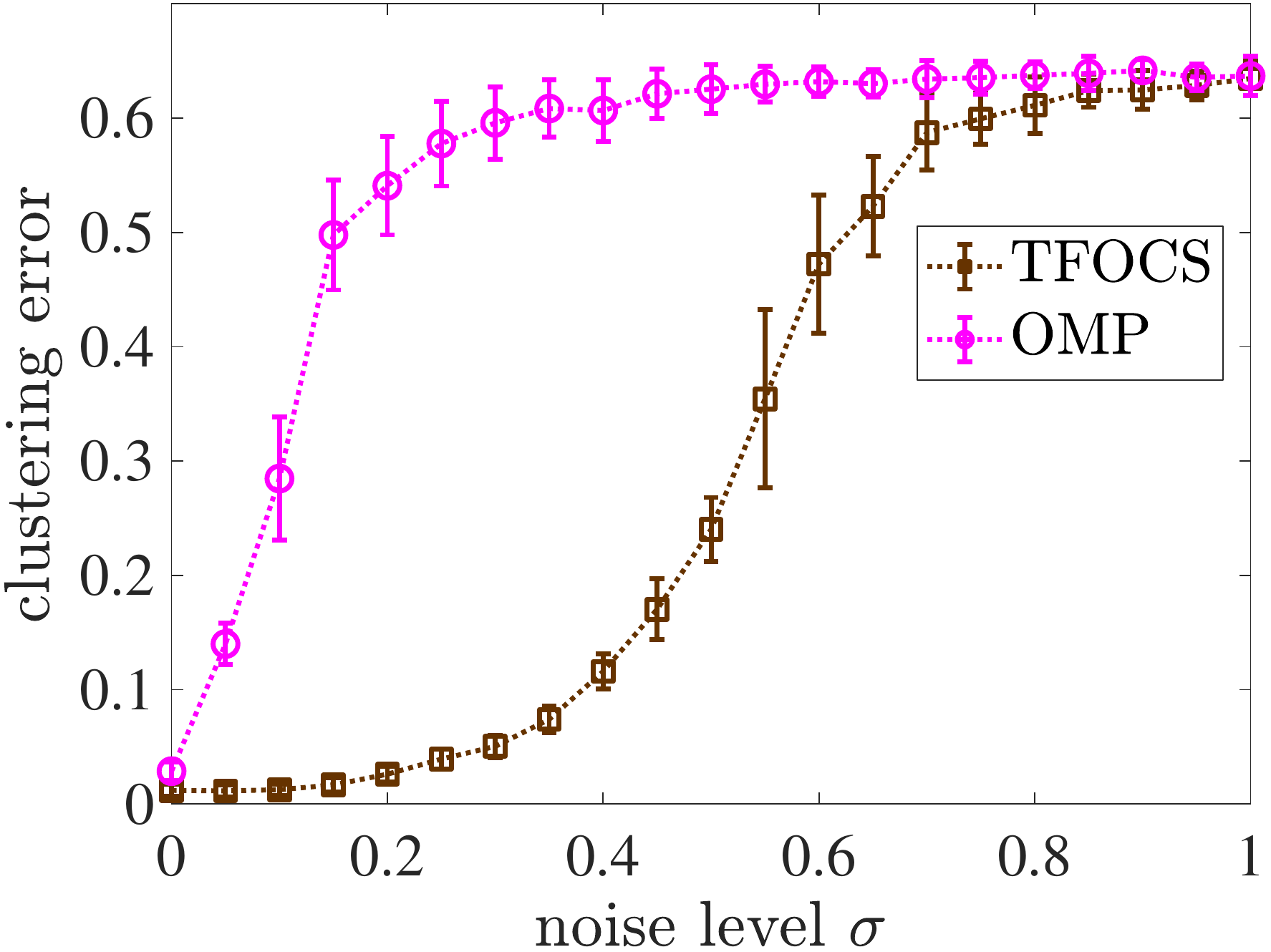}
	}    
	
	\caption{Clustering error of SSC-$\ell_0$ on synthetic data for varying $\sigma$ (noise level). \label{fig:syn-l0}
	}
\end{figure*}

The clustering error results, showing average and standard deviation over $20$ independent trials, 
for two choices of sparsity $k=10$ and $k=20$, are plotted in Figure \ref{fig:syn-l0},
where $k$ is the sparsity parameter in Eq.~\eqref{eq:SSCl0}. 
As expected, larger values of the noise level $\sigma$ result in lower accuracy clustering results. However, we see that our TFOCS solver consistently outperforms OMP 
for both  
$k=10$ and $k=20$, and the effect is more pronounced for $k=20$. Since each subspace in this example is $10$-dimensional, it is worth pointing that the proposed TFOCS solver is less sensitive to the choice of sparsity $k$ than OMP.

\begin{figure}
	\centering
	\includegraphics[width=0.49\textwidth]{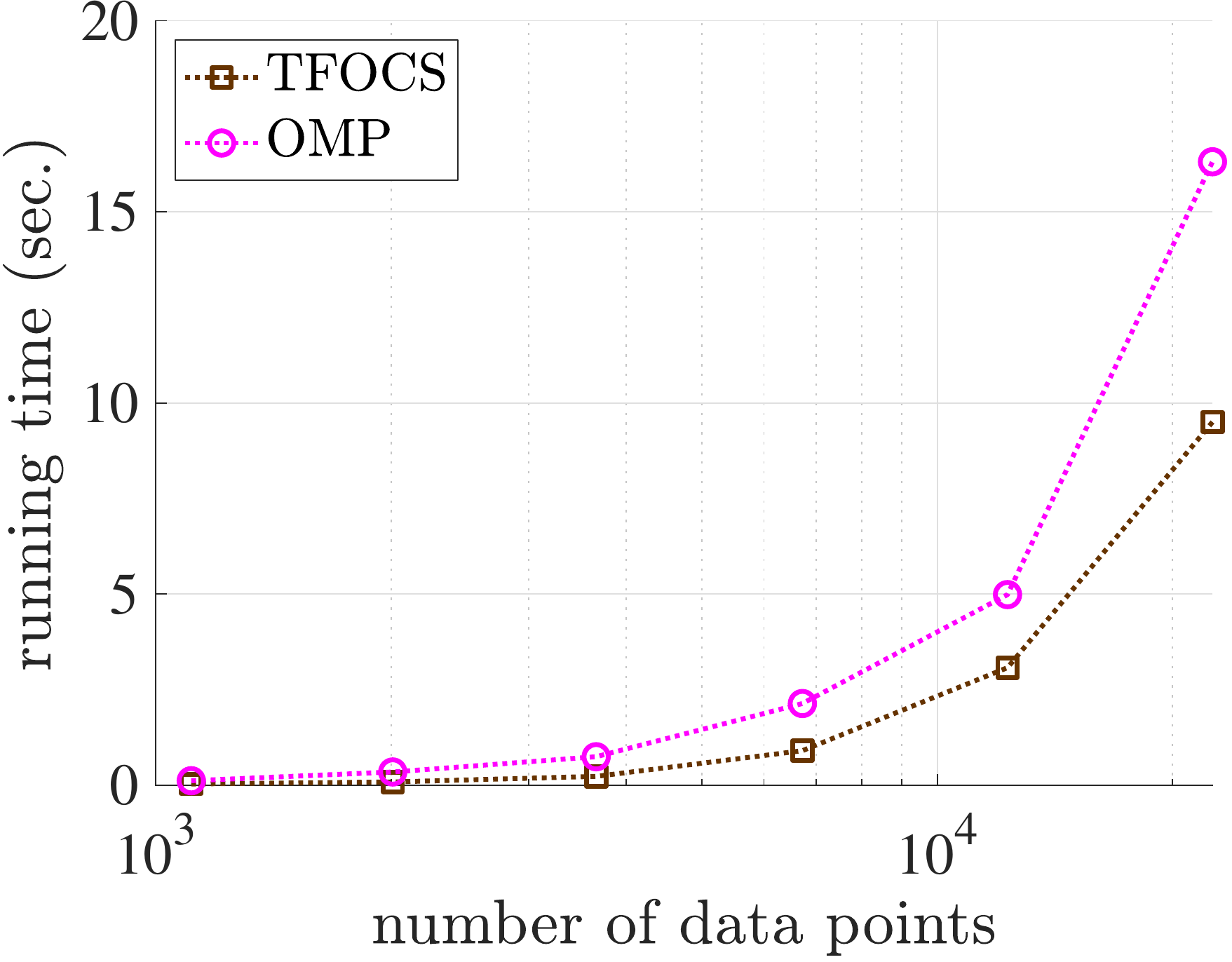}
	\includegraphics[width=0.49\textwidth]{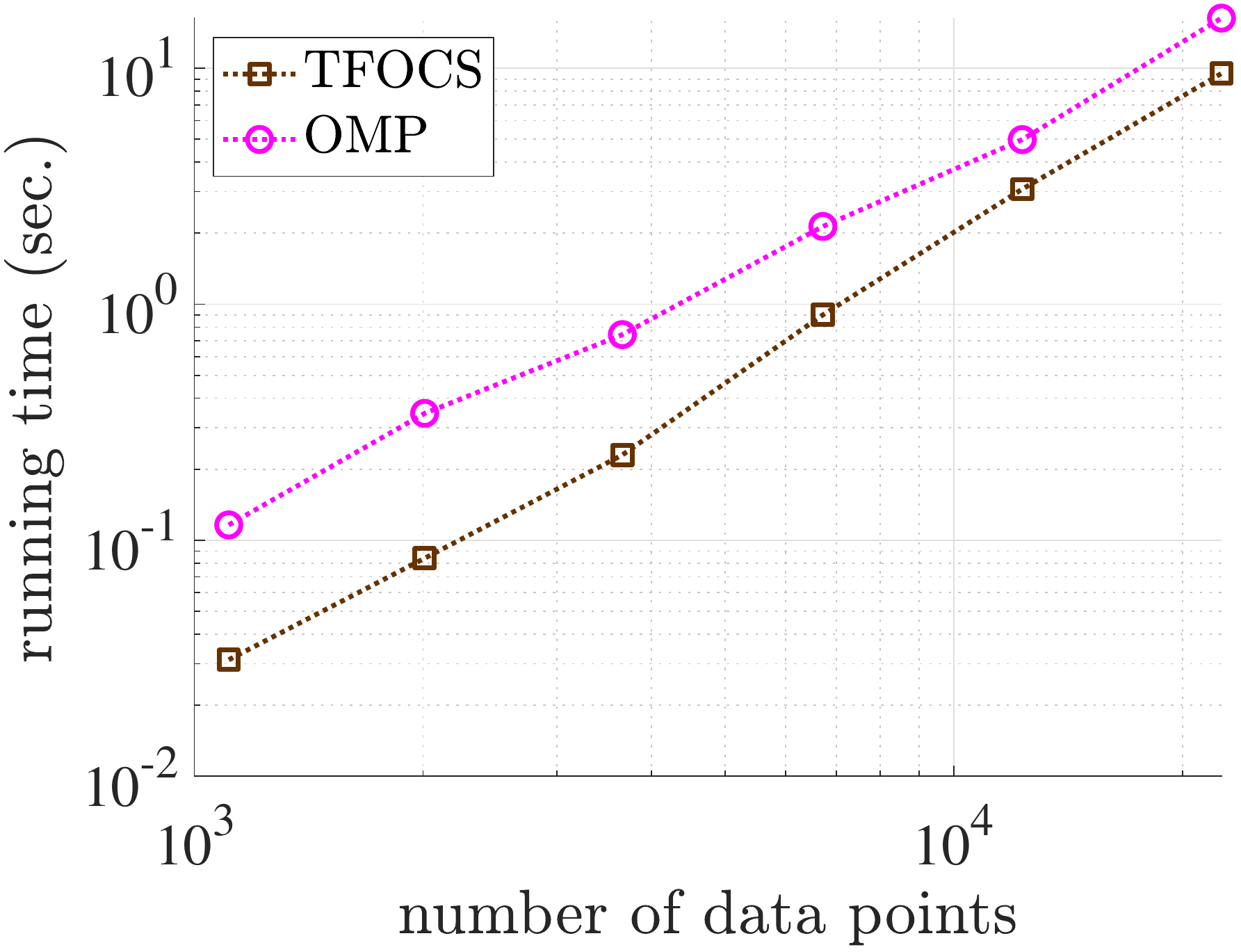}
	\caption{Running time of SSC-$\ell_0$ on synthetic data for varying number of data points $n$.
		Left: time on a linear scale. Right: same data, but time on a logarithmic scale. }
	\label{fig:syn-l0-time}
\end{figure}

To compare the efficiency of our proximal gradient solver for the SSC-$\ell_0$ problem with OMP, the running times required to achieve a certain level of accuracy for various number of data points from $n=600$  to $n=22,\!500$ 
are plotted in Figure \ref{fig:syn-l0-time} (other parameters such as the dimension of subspaces and the ambient dimension are unchanged). To be more specific, we run OMP for $10$ iterations and then run our TFOCS solver to match the clustering error produced by OMP, and report the corresponding running time. In this experiment, it is observed that TFOCS is faster to reach OMP's accuracy. 

To summarize, our proximal SSC-$\ell_0$  algorithm  is significantly more accurate than OMP when the noise is high and/or $k$ is over-specified. Furthermore, our solver is the only known algorithm to solve the affine space variant of  SSC-$\ell_0$.

\section{Conclusion}\label{sec:conc}
We proposed two efficient proximal gradient methods for finding sparse representation vectors of data points that lie in or close to a union of affine subspaces. We also presented a detailed performance and complexity analysis of our proximal solvers. In addition, an efficient implementation of the popular ADMM technique for solving $\ell_1$ norm regularized SSC optimization problems is provided. Overall, the two proposed proximal solvers and our implementation of ADMM substantially reduces the computational cost of solving large-scale SSC optimization problems. A key advantage of our proximal solver for SSC-$\ell_1$ is the lack of additional parameter tuning, which makes it much more efficient than ADMM (if one does cross-validation to find the correct parameter). Experimentally, ADMM does appear to be sensitive to its additional parameter $\rho$. Finally, our proposed proximal solver for SSC-$\ell_0$ has the ability to directly deal with the more general case of affine subspaces, and experimentally it appears to be less sensitive to the choice of sparsity parameter compared to the existing algorithm that uses OMP. As a final note, it is worth pointing out  our proposed solvers can be  adopted in a recent line of work, e.g., \cite{you2018scalable}, that uses exemplars or representative points to further achieve scalability to large data sets.

\bibliographystyle{plain}
\bibliography{phd_farhad}

\end{document}